\documentclass[10pt,twocolumn,letterpaper]{article}

\usepackage{wacv}              

\definecolor{wacvblue}{rgb}{0.21,0.49,0.74}

\def\wacvPaperID{945} 
\def\confName{WACV}
\def\confYear{2027}

\title{
Guardian: Detecting Robotic Manipulation Failures  with Vision-Language Models
}

\author{Paul Pacaud \quad Ricardo Garcia \quad Shizhe Chen \quad  Cordelia Schmid \\
Inria, École normale supérieure, CNRS, PSL Research University\\
{\tt\small firstname.lastname@inria.fr}
}

\usepackage{graphicx}	
\usepackage{amsmath}	
\usepackage{amssymb}	
\usepackage{arydshln}
\usepackage{booktabs}
\usepackage{times}
\usepackage{microtype}
\usepackage{epsfig}
\usepackage{caption}
\usepackage{float}
\usepackage{placeins}
\usepackage{color, colortbl}
\usepackage{stfloats}
\usepackage{enumitem}
\usepackage{tabularx}
\usepackage{xstring}
\usepackage{multirow}
\usepackage{xspace}
\usepackage{url}
\usepackage{subcaption}
\usepackage{xcolor}
\usepackage{tcolorbox}
\usepackage[hang,flushmargin]{footmisc}
\usepackage{wrapfig} 
\usepackage{adjustbox}
\usepackage{pifont}
\usepackage{array}
\usepackage{makecell}
\usepackage[pagebackref,breaklinks,colorlinks,allcolors=wacvblue]{hyperref}
\usepackage[capitalize,nameinlink]{cleveref}

\crefname{figure}{Fig.}{Figs.}
\crefname{table}{Table}{Tables}
\crefname{section}{Sec.}{Secs.}
\crefname{equation}{Eq.}{Eqs.}
% ...

\newcommand{\R}[1]{{%
    \textbf{%
        \ifstrequal{#1}{1}{\textcolor{red}{R#1}}{%
        \ifstrequal{#1}{2}{\textcolor{blue}{R#1}}{%
        \ifstrequal{#1}{3}{\textcolor{magenta}{R#1}}{%
        \ifstrequal{#1}{4}{\textcolor{teal}{R#1}}{%
                           \textcolor{cyan}{R#1}%
        }}}}%
    }%
}}

\begin{document}
\maketitle
\begin{abstract}
Robust robotic manipulation requires reliable failure detection and recovery. Although recent Vision-Language Models (VLMs) show promise in robot failure detection, their generalization is severely limited by the scarcity and narrow coverage of failure data. 
To address this bottleneck, we propose an automatic framework for generating diverse robotic planning and execution failures across both simulated and real-world environments.
Our approach perturbs successful manipulation trajectories to synthesize failures that reflect realistic failure distributions, and leverages VLMs to produce structured step-by-step reasoning traces.
This yields GuardianFail-36k, a large-scale failure reasoning dataset built upon the RLBench simulator and the BridgeDataV2 real-robot dataset. 
Using GuardianFail-36k, we train Guardian, a multi-view reasoning VLM for unified planning and execution verification.
Guardian achieves state-of-the-art performance on three unseen real-world benchmarks: RoboFail, RoboVQA, and our newly introduced UR5-Fail. When integrated with a state-of-the-art LLM-based manipulation policy, it consistently boosts task success rates in both simulation and real-world deployment. These results demonstrate that scaling high-quality failure reasoning data is critical for improving generalization in robotic failure detection. Code, Data, and Models are available at \href{https://www.di.ens.fr/willow/research/guardian/}{\textit{https://www.di.ens.fr/willow/research/guardian/}}.
\end{abstract}

\section{Introduction}
\label{sec:intro}


\begin{figure}[t]
    \centering
    \includegraphics[width=\linewidth]{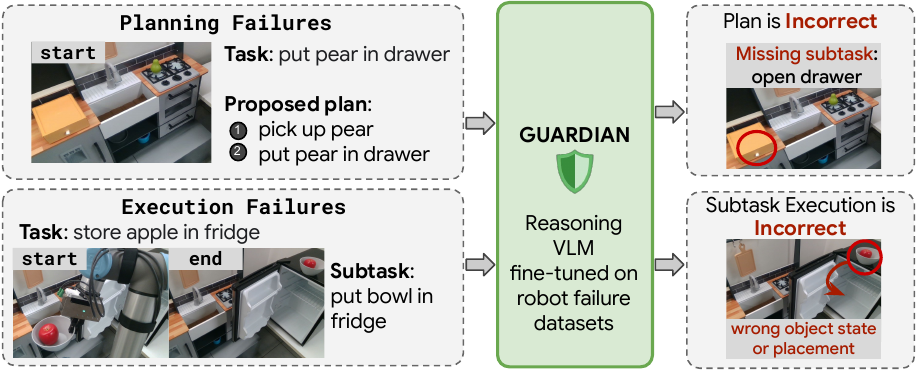}
    \caption{Illustration of our Guardian model - a VLM fine-tuned on our constructed failure datasets. It detects planning failures (top) and execution failures (bottom) in robotic manipulation.}
    \vspace{-1em}
    \label{fig:teaser}
\end{figure}

Advances in Large Language Models (LLMs)~\cite{mistralsmall,grattafiori2024llama3herdmodels} and Vision-Language Models (VLMs)~\cite{zhu2025internvl3exploringadvancedtraining} have significantly improved vision-language robotic manipulation. 
Yet, existing models remain vulnerable to diverse failures~\cite{sinha2023systemlevelviewoutofdistributiondata,Kawaharazuka_2024,kroemer2020reviewrobotlearningmanipulation} such as incorrect task decomposition, object confusion, or unstable grasps, which compound over time and degrade reliability.
As a result, automatic failure detection and recovery have received growing attention~\cite{liu2023reflect,chen2024automatingrobotfailurerecovery,duan2025aha,agia2024unpackingfailuremodesgenerative,etukuru2024robotutilitymodelsgeneral,ifailsense2026,zeng2025vifailback}.

Leveraging their strong generalization ability, LLMs and VLMs have been increasingly explored for failure detection.
Some methods~\cite{liu2023reflect,etukuru2024robotutilitymodelsgeneral,duan2024manipulateanythingautomatingrealworldrobots}, directly prompt pretrained foundation models to detect failures, optionally enhanced with chain-of-thought (CoT) reasoning~\cite{agia2024unpackingfailuremodesgenerative,nvidia2026cosmosreason2,cot2022} or multi-agent code generation~\cite{zhou2024code}.
While promising, these approaches suffer from a large domain gap: robotic observations differ substantially from web-scale pretraining data, and accurate failure detection requires fine-grained, embodied reasoning beyond generic visual understanding.
Therefore, recent work~\cite{duan2025aha,ifailsense2026,robofac2025} has shifted toward fine-tuning VLMs on robot failure datasets to better bridge the gap.

A fundamental bottleneck, however, is the scarcity of large-scale, high-quality failure data. Most robot datasets predominantly contain successful demonstrations~\cite{khazatsky2025droidlargescaleinthewildrobot,embodimentcollaboration2024openxembodimentroboticlearning,pumacay2024colosseumbenchmarkevaluatinggeneralization}, providing limited failure examples. Collecting failures by rolling out policies is time-consuming and unsafe, while manual curation~\cite{chen2024automatingrobotfailurerecovery,bu2025agibot} is labor-intensive and typically lacks diversity. Several prior approaches~\cite{duan2025aha,agia2024unpackingfailuremodesgenerative,dai2024racer} rely on simulated failure examples, but these suffer from sim-to-real gap~\cite{simtorealgapzhao2020}, and provide limited coverage of both low-level execution errors and high-level planning failures~\cite{ifailsense2026}. 

\begin{figure*}[ht]
    \vspace{0.6em}
    \centering
    \includegraphics[width=0.95\linewidth]{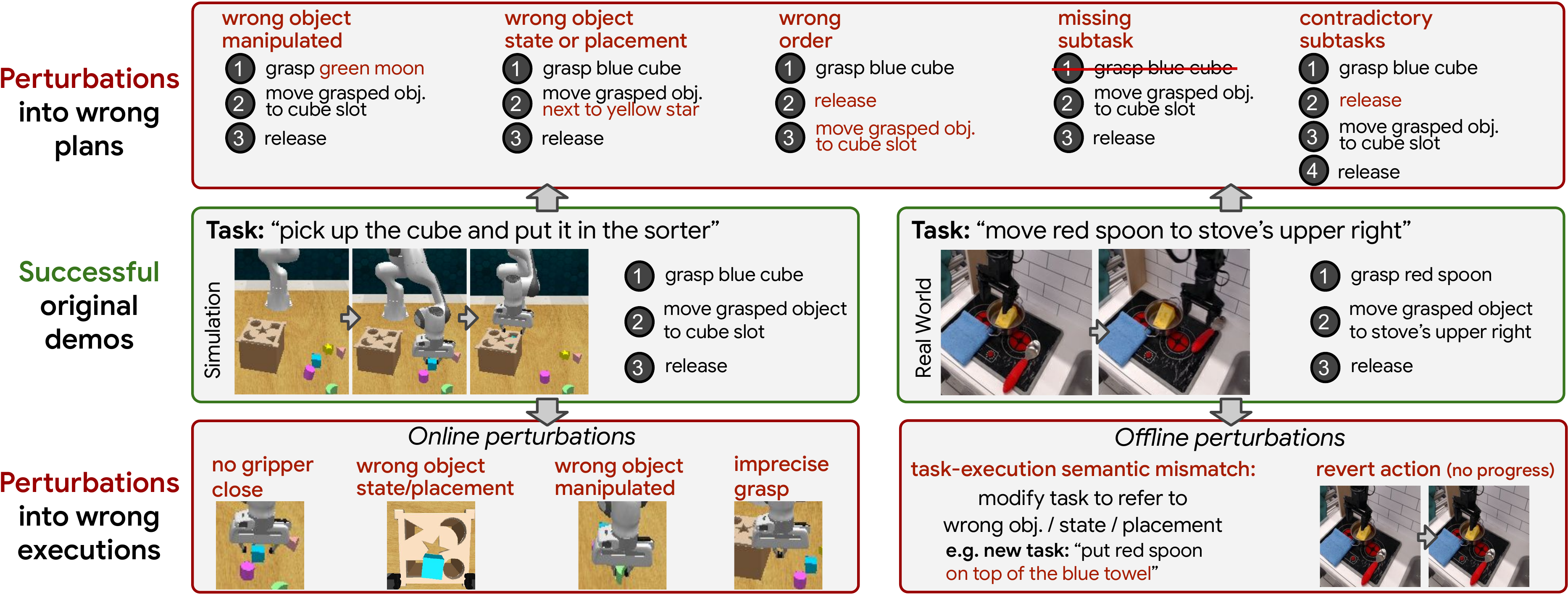}
    \caption{Failure Data Generation Framework. We construct failure cases both online in simulation (RLBench, left), and offline on the real-world dataset (BridgeDataV2, right). For each positive example, given its correct plan and successful trajectory, we generate a corresponding incorrect plan and unsuccessful trajectory.}
    \label{fig:guardian_data_pipeline}
    \vspace{-1.3em}
\end{figure*}

To address these limitations, we propose an automatic failure generation framework that synthesizes diverse planning and execution failures across simulated and real-world environments.
Starting from successful demonstrations, we procedurally perturb task plans and subtask executions to create realistic failures, augmenting each example with structured step-by-step reasoning traces. 
This enables the construction of \emph{GuardianFail-36k}, a large-scale failure reasoning dataset containing over 36K examples. It includes \emph{RLBench-Fail} built using the RLBench simulator~\cite{james2019rlbenchrobotlearningbenchmark} and \emph{BridgeDataV2-Fail} derived from the BridgeDataV2 real-robot dataset~\cite{walke2024bridgedatav2datasetrobot}, see Figure~\ref{fig:guardian_data_pipeline}.
GuardianFail-36k provides balanced success and failure samples, multi-view visual observations, and explicit CoT supervision for plan and subtask-level verification. 

Building on \emph{GuardianFail-36k}, we develop \emph{Guardian}, a multi-view reasoning VLM fine-tuned for unified planning and execution failure detection. Guardian follows a visual question answering problem: conditioned on task instructions, proposed plans or subtasks, and multi-view observations, it produces explicit reasoning traces to predict failures.
To further support realistic evaluation, we introduce a new real-robot benchmark \emph{UR5-Fail} constructed using the same failure generation framework. 
Guardian achieves state-of-the-art performance on three unseen real-world failure benchmarks, namely \emph{RoboFail}~\cite{liu2023reflect},  \emph{RoboVQA}~\cite{sermanet2024} and \emph{UR5-Fail}.
When integrated as a plug-and-play verification module into a LLM-based manipulation system, Guardian improves task success in both simulation and real-robot experiments.
Extensive ablations demonstrate the benefits of scaling structured, cross-environment failure reasoning data.

\noindent In summary, our contributions are three-fold:
\begin{itemize}[leftmargin=*, topsep=0em]
    \item We propose an automatic cross-environment failure synthesis framework that generates diverse planning and execution errors with structured reasoning supervision, yielding the large-scale robot failure dataset \emph{GuardianFail-36k}.
    \item We develop \emph{Guardian}, a multi-view reasoning VLM fine-tuned on the GuardianFail-36k dataset for unified planning and execution failure detection.
    \item We show that scaling structured failure reasoning data yields state-of-the-art detection performance and improves task success when deployed as a plug-and-play verifier.
\end{itemize}
We will release datasets, code, and models.

\section{Related Work}
\label{sec:related-work}

\noindent\textbf{Vision-Language Robotic Manipulation.}
Foundation models have enabled progress in vision-language robotic manipulation~\cite{zhu2025internvl3exploringadvancedtraining,openai2024gpt4ocard}, including end-to-end VLA policies such as $\pi_0$~\cite{black2024pi} and 3D-based approaches for improved spatial reasoning~\cite{garcia2025generalizablevisionlanguageroboticmanipulation,goyal2023rvtroboticviewtransformer}. Notably, 3D-LOTUS++~\cite{garcia2025generalizablevisionlanguageroboticmanipulation} achieves state-of-the-art through a modular design that combines LLM-based task planning and 3D-based execution policies.
Yet, robust robotic manipulation remains challenging, as planning and execution errors can accumulate~\cite{wu2025robomind}. 
To improve robustness, recent approaches incorporate runtime failure monitoring to trigger correction or retry~\cite{duan2024manipulateanythingautomatingrealworldrobots,etukuru2024robotutilitymodelsgeneral,dai2024racer,Zhao_2026_CVPR,FPCVLA_YANG2026131742,ma2026cyclevla,chong2026robusttaskplanning}.
We build on this by scaling structured failure-reasoning data and integrating failure verification with manipulation policies.

\noindent\textbf{Robotic Failure Detection Methods.}
Early rule-based failure detection methods~\cite{de1998execution,gianni2011unified} struggle to generalize beyond predefined task structures. Learning-based approaches address this limitation and can be broadly categorized based on whether they require robot failure data for training.

\emph{Training-free methods.}
One line of work performs out-of-distribution (OOD) detection~\cite{canwedetectfailurewithoutfailuredata2025} or temporal inconsistency detection~\cite{agia2024unpackingfailuremodesgenerative} over internal policy representations to flag failures without failure supervision. Another line prompts LLMs or VLMs for failure assessment~\cite{ahmad2025unifiedframework,hosseinzadeh2025kite,guo2024doremi} using techniques such as hierarchical CoT reasoning~\cite{liu2023reflect,agia2024unpackingfailuremodesgenerative} or constraint-aware visual programming~\cite{zhou2024code}. 
While these approaches avoid collecting data, they rely on policy-specific signals or prompt engineering, and cannot learn from failure data.

\emph{Training-based methods.}
Recent works~\cite{duan2025aha,ifailsense2026,nvidia2026cosmosreason2,gu2025safe,pmlr-v232-du23b,armor2026,lin2025failsafe} fine-tune VLMs as failure detectors using annotated trajectories, but mostly target execution-stage verification. AHA~\cite{duan2025aha} and I-Fail-Sense~\cite{ifailsense2026} compress multi-view inputs into a single concatenated image for detecting subtask-level execution failures or instruction–behavior misalignment. 
SAFE~\cite{gu2025safe} relies on internal policy representation to train a failure classifier. 
Cosmos-Reason~\cite{nvidia2026cosmosreason2} addresses general embodied reasoning trained with supervised and reinforcement learning, where failure detection is one downstream task. 
ARMOR~\cite{armor2026} uses multi-round self-refinement reasoning, but trains only on execution failures and requires multiple inference passes per sample. 
Compared to prior methods, our Guardian model leverages large-scale failure reasoning data to enable multi-view, explicit reasoning for unified planning and execution verification.

\noindent\textbf{Robot Failure Datasets.}
Collecting real-world robot failures at scale is challenging: policy rollouts are time-consuming, unsafe, and require manual annotation. RoboFail~\cite{liu2023reflect} provides a hand-crafted dataset spanning simulation and real-world, but covers limited tasks and failures. ViFailback~\cite{zeng2025vifailback} focuses on single-view, single-embodiment real-world diagnosis and requires substantial teleoperation.

To reduce collection cost, several works rely on synthetic failure generation. Sentinel~\cite{agia2024unpackingfailuremodesgenerative} induces failures via out-of-distribution rollouts but covers only four tasks. SAFE~\cite{gu2025safe} utilizes the final sparse reward for policy rollouts and lacks dense failure supervision for each step.
AHA~\cite{duan2025aha} perturbs trajectories in RLBench~\cite{james2019rlbenchrobotlearningbenchmark}, generating purely simulated data, excludes high-level planning failures, and the dataset has not been publicly released. RoboFAC~\cite{robofac2025} adds reasoning labels in simulation, with only a limited real-world subset via teleoperation. 
I-Fail-Sense~\cite{ifailsense2026} synthesizes failures from RLBench and DROID, but focuses on semantic mismatches, leaving planning and low-level control errors underexplored. Another line instead synthesizes robot data with video-generation models~\cite{black2023susie,guo2025ctrl,ye2026anchordream}, but targets mainly successful demos rather than failures, and often lacks physical plausibility~\cite{deng2026rethinking, jiang2026robowmbench}. In contrast, we propose an automated pipeline that generates diverse planning and execution failures across both simulation and real robots, producing realistic failure modes at scale with multi-view observations and fine-grained, step-by-step reasoning supervision.

\section{GuardianFail-36k: Cross-Environment Robot Failure Reasoning Datasets}
\label{sec:dataset}

\subsection{Data Sources}
\label{sec:data_source}

Simulated data enables controlled failure generation through procedural perturbations~\cite{duan2025aha}, while real robot data reduces the sim-to-real gap but requires human effort~\cite{liu2023reflect}. 
To balance precise control and real-world fidelity, we use both simulated and real-robot data to construct robot failure datasets.
We propose an automated method, illustrated \cref{fig:guardian_data_pipeline}, that derives planning and execution failures directly from successful demonstrations, avoiding manual collection. 
Tasks are decomposed into subtasks with corresponding video segments, which form the basis for generating failures.

\textbf{Simulated Data}. 
We use the RLBench~\cite{james2019rlbenchrobotlearningbenchmark} simulator, selecting 52 tasks from PerAct~\cite{shridhar2022perceiveractormultitasktransformerrobotic} and GemBench~\cite{garcia2025generalizablevisionlanguageroboticmanipulation} benchmarks. For each task, we generate successful scripted trajectories and segment them into subtasks following~\cite{garcia2025generalizablevisionlanguageroboticmanipulation}. 

\textbf{Real Robot Data.} We use BridgeDataV2~\cite{walke2024bridgedatav2datasetrobot} with 
ECoT annotations ~\cite{zawalski2025roboticcontrolembodiedchainofthought}, which provide subtasks and object labels using VLMs.
We further clean these annotations automatically using heuristics and Mistral-Small-3.1-24B~\cite{mistralsmall} to filter episodes with missing targets or unreliable bounding boxes.
We increase the number of successful samples by reversing successful executions, by swapping their start and end images, and updating the associated instructions accordingly (e.g., ``open drawer" becomes ``close drawer"). This yields around 20\% additional successful demonstrations.

\subsection{Automated Failure Data Generation}
\label{sec:dataset_gen}

We design failure modes based on established failure taxonomies~\cite{liu2023reflect,duan2025aha} and analysis of robot policy failures~\cite{wu2025robomind}.
The failures are categorized into two types: planning and execution.
A planning error denotes an incorrect decomposition of a task into subplans, whereas an execution error reflects unsuccessful completion of a subplan.

\noindent \textbf{Planning Failures.} 
Each planning example comprises the task instruction, plan, and the initial front-view image. As shown in \cref{fig:guardian_data_pipeline} (top), we craft five  planning failure types:

\begin{enumerate}[label=(\arabic*), leftmargin=*, itemsep=-0.2em, topsep=0em]
    \item \emph{Wrong object manipulated} – some subtasks manipulate the wrong object.
    \item \emph{Wrong object state or placement} – some subtasks select the wrong target location, or state for the correct object.
    \item \emph{Wrong order} – one or several subtasks are not in the correct order, violating causal dependencies.
    \item \emph{Missing subtask} – required subtasks are missing from the plan, breaking task completeness.
    \item \emph{Contradictory subtasks} – some subtasks conflict with each other.
\end{enumerate}
Types 1–2 use an LLM (Mistral-Small-24B): conditioned on the instruction, plan, and visible objects, it alters a single element of the plan, one manipulated object, or one target location/state, keeping all other subtasks fixed.
Types 3-5 are created through rule-based perturbations by swapping, deleting, or inserting. 

\noindent \textbf{Execution Failures.} 
Each execution example contains the task and subtask descriptions, plus pre-/post-action multi-view images. In simulation, we directly perturb subtask-level actions (\cref{fig:guardian_data_pipeline}, bottom left), leveraging the simulator’s precise control. A randomly selected subtask on the trajectory is modified using four failure modes: 

\begin{enumerate}[label=(\arabic*), leftmargin=*, itemsep=-0.1em, topsep=0em]
    \item \emph{No gripper close} – the gripper is correctly positioned to grasp the object, but it fails to close its jaws.
    \item \emph{Wrong object state or placement} – the correct object is manipulated but ends in an incorrect state or placement.
    \item \emph{Wrong object manipulated} – the wrong object is used.
    \item \emph{Imprecise grasping/pushing} – the gripper attempts to grasp or push the correct object by moving toward it and closing its jaws, but misses due to inaccurate positioning.
\end{enumerate}

For real robot data, failures must be derived from already-collected successful demonstrations. Synthesizing perturbed executions with image-editing is a natural alternative, but current models remain physically unreliable~\cite{deng2026rethinking, jiang2026robowmbench}. We therefore preserve the recorded trajectory unaltered and perturb only its paired subtask text instruction (\cref{fig:guardian_data_pipeline}, bottom right), constructing a known failure mode:

\begin{enumerate}[label=(\arabic*), leftmargin=*, itemsep=-0.15em, topsep=0em]
    \item \emph{Task-execution semantic mismatch} — an LLM (prompted with the original instruction and visible objects), or a rule-based preposition swap, generates a semantically altered instruction while preserving the start/end images.
    \item \emph{Revert action} — keep the instruction unchanged; replace the end image with the start one to show no progress.
\end{enumerate}

\begin{figure}[t]
    \centering
    \includegraphics[width=\linewidth]{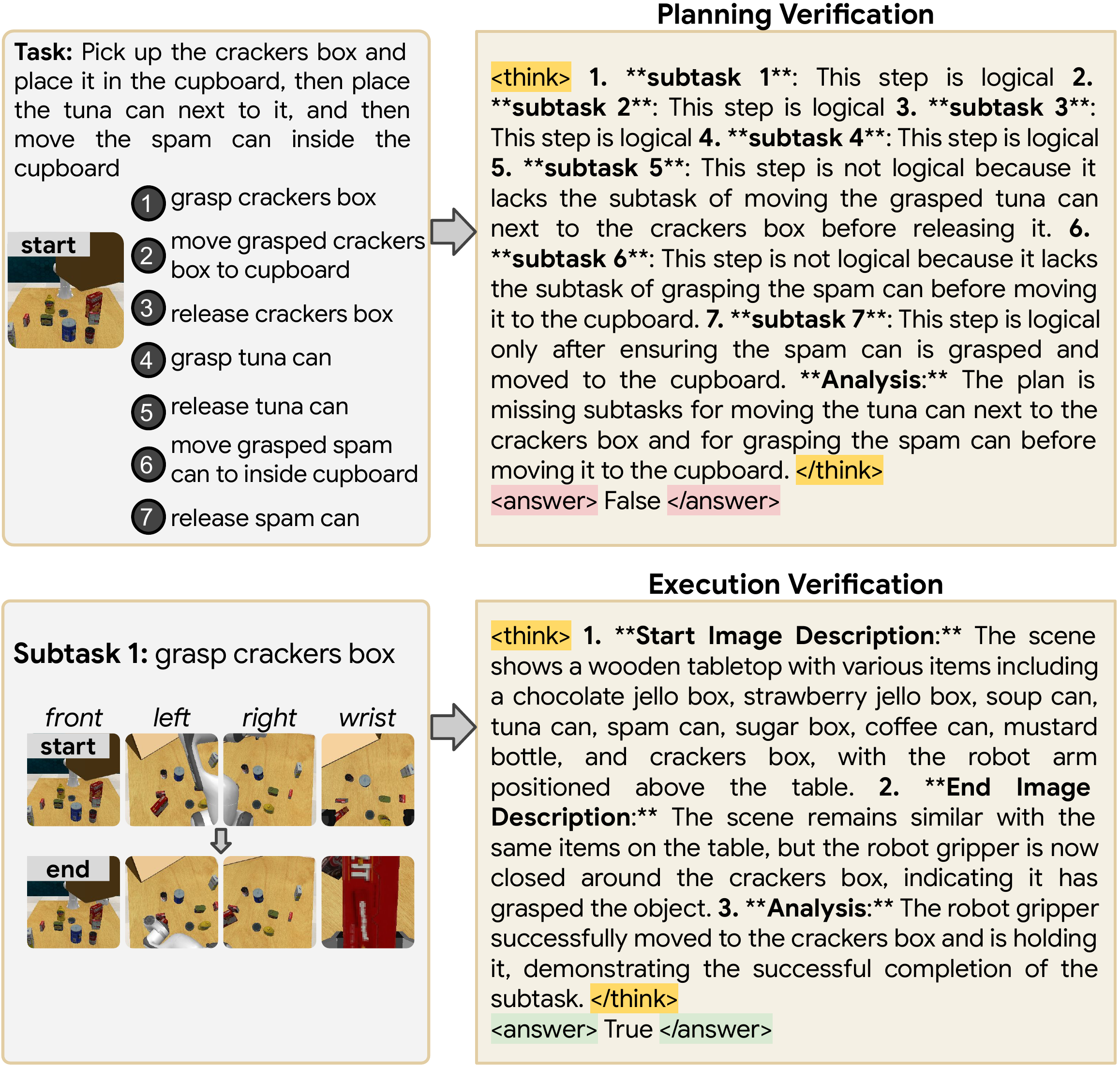}
    \caption{Planning and execution examples with CoT.}
    \label{fig:cot_examples}
    \vspace{-1em}
\end{figure}

\subsection{Chain-of-Thought (CoT) Generation}
\label{sec:cot_generation}

CoT reasoning has shown promise in improving the interpretability and performance of VLMs~\cite{zhang2024improve}, yet it is lacking from existing robot failure generation methods such as AHA or I-Fail-Sense~\cite{duan2025aha, ifailsense2026}.
Therefore, we further explore whether reasoning can help failure detection. 
We introduce an automatic method to generate step-by-step CoTs for training reasoning models.
For each sample, we first collect the object category, spatial location, and robot state from the RLBench simulator or from ECoT~\cite{zawalski2025roboticcontrolembodiedchainofthought} annotations, together with the corresponding failure reason.
We then prompt a large reasoning-capable VLM (InternVL3-38B)~\cite{zhu2025internvl3exploringadvancedtraining} to generate step-by-step reasoning traces based on the initial text–image inputs and the aforementioned information.
For planning samples, the model is instructed to sequentially verify each subtask and subsequently analyze the overall plan.
For execution samples, the model is guided to describe the pre- and post-action images before assessing subtask completion.
The reasoning trace contains 118 tokens on average.
\cref{fig:cot_examples} illustrates training examples with CoTs.

\begin{table}[t]
\vspace{0.4em}
\centering
\caption{
Statistics of the GuardianFail-36k dataset and three real-world robot failure detection benchmarks. 
Our constructed datasets (GuardianFail and UR5-Fail) contain balanced success and failure cases covering both execution and planning errors. 
}
\label{tab:dataset_stats}
\resizebox{\columnwidth}{!}{
\begin{tabular}{l c cccccc}
    \toprule
    & & \multicolumn{2}{c}{\textbf{Training}} & \multicolumn{2}{c}{\textbf{Validation}} & \multicolumn{2}{c}{\textbf{Test}} \\
    \cmidrule(lr){3-4} \cmidrule(lr){5-6} \cmidrule(lr){7-8}
    \textbf{Dataset} & \textbf{Env.} & Exec & Plan & Exec & Plan & Exec & Plan \\
    \midrule
    \multicolumn{8}{l}{\textbf{GuardianFail (Ours)}} \\
    \quad RLBench-Fail & Sim & 12358 & 5808 & 1000 & 500 & 1000 & 500 \\
    \quad BridgeDataV2-Fail & Real & 7830 & 4880 & 1000 & 500 & 1000 & 500 \\
    \midrule
    \multicolumn{8}{l}{\textbf{Real-World Robot Failure Detection Benchmarks}} \\

    UR5-Fail (Ours) & Real & - & - & - & - & 140 & 140 \\

    RoboFail~\cite{liu2023reflect} & Real & - & - & - & - & 153 & 30 \\
    RoboVQA~\cite{sermanet2024} & Real & - & - & - & - & 357 & - \\
    \bottomrule
\end{tabular}
}
\vspace{-1em}
\end{table}
\begin{figure}[t]
    \vspace{0.8em}
    \centering
    \includegraphics[width=\linewidth]{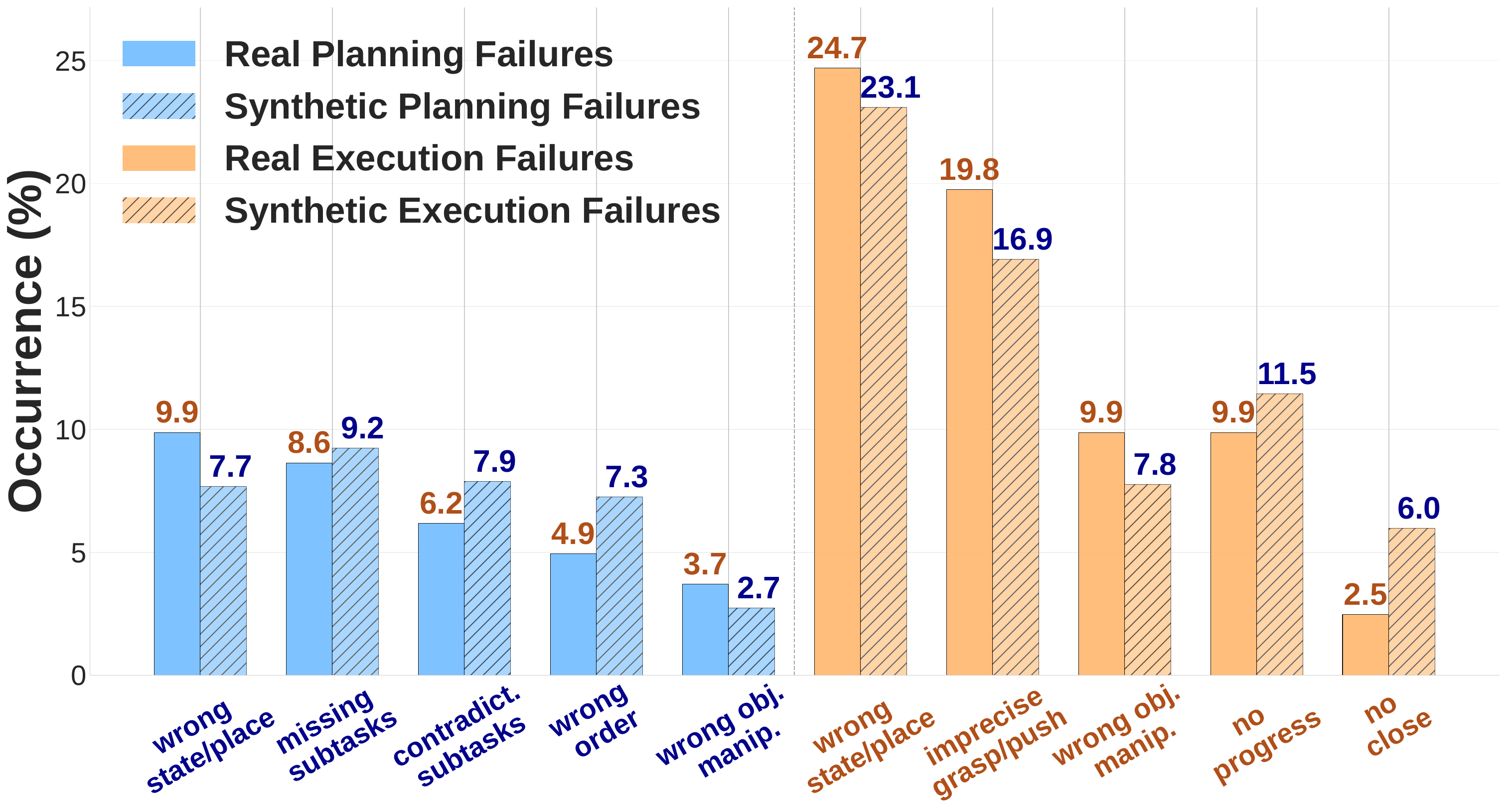}
    \caption{Failure mode distributions in real executions and our constructed data.}
    \label{fig:dataset_dist_cmpr}
    \vspace{-1em}
\end{figure}
\subsection{Real-Robot, Policy-Driven Data Collection}
\label{sec:dataset_ur5}

To further support realistic evaluation, we curate \emph{UR5-Fail}, a real-robot dataset, collected using a UR5 arm with three cameras.
We run the 3D-LOTUS++ policy~\cite{garcia2025generalizablevisionlanguageroboticmanipulation} on 16 unique tasks, 
recording initial and final multi-view images for each subtask. Subtasks are manually labeled as success or failure to obtain execution failure data. For planning failures, we annotate ground-truth plans and generate failures using the method described in \cref{sec:dataset_gen}. Unlike \emph{RoboFail}~\cite{liu2023reflect}, which is single-view and relies solely on teleoperation, \emph{UR5-Fail} is three-view and features autonomous policy rollouts yielding more realistic failures. 

\subsection{Dataset Statistics and Evaluation}
\label{sec:dataset_summary}

GuardianFail-36k (\emph{RLBench-Fail}, \emph{BridgeDataV2-Fail}) and \emph{UR5-Fail}, contain balanced success/failure examples across both planning and execution, with reasoning traces.
GuardianFail-36k is split into training, validation, and test sets, with the validation and test sets featuring unseen tasks/environments to evaluate generalization, see \cref{tab:dataset_stats} top.
\cref{tab:dataset_stats} bottom compares \emph{UR5-Fail} with two existing real-world datasets and shows a more balanced distribution between execution and planning. 

To measure the quality and diversity of our synthetic datasets, i.e., whether the generated failures reflect real policy execution, we run the 3D-LOTUS++ policy~\cite{garcia2025generalizablevisionlanguageroboticmanipulation} on 92 RLBench tasks and manually annotate failure modes for 3 failure episodes per task.
As shown in \cref{fig:dataset_dist_cmpr}, our designed failure modes reflect real failures, and the overall distribution of our synthetic and real failures remains similar.

\begin{figure*}[t]
    \vspace{0.8em}
    \centering
    \includegraphics[width=0.97\linewidth]{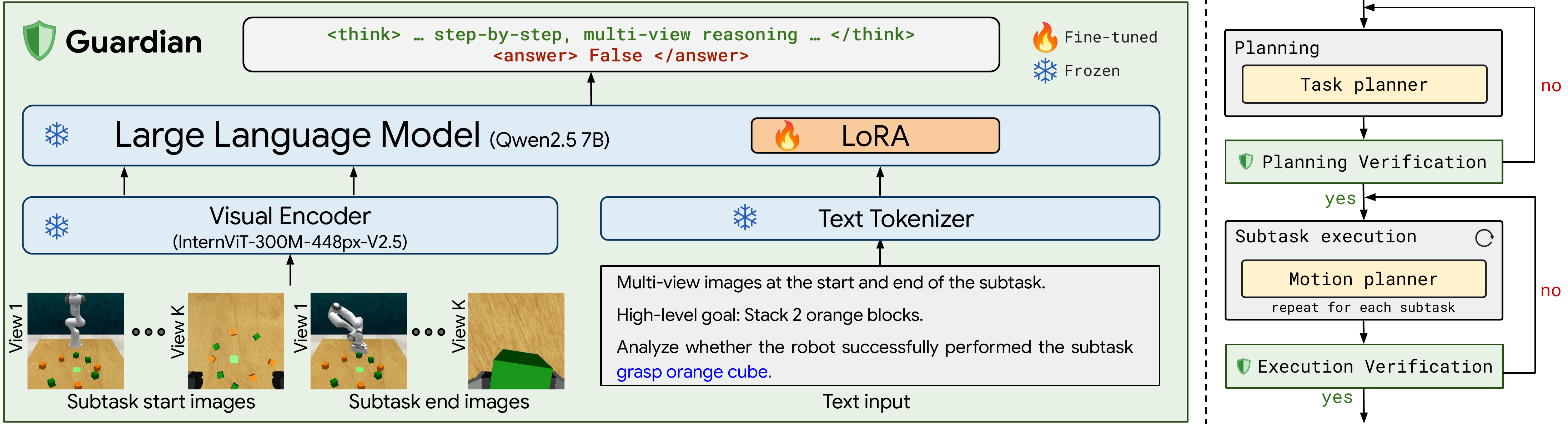}
    \caption{Left: Overview of the Guardian model architecture. Right: Integration of Guardian model into a robot manipulation pipeline for planning and execution verification. The failure reasoning trace is reinjected to help with replanning.
    } 
    \label{fig:model_architecture}
    \vspace{-1.3em}
\end{figure*}
\section{Guardian: A multi-view reasoning VLM for robot failure detection}
\label{sec:method}

\subsection{Problem Formulation} 

We formulate robot failure detection as a visual question answering problem. For planning verification, given a high-level task instruction $T$, a proposed plan $P = (P_1, \cdots, P_N)$, and the initial visual context $I_{\text{start}}$, the model $\text{VLM}_{\text{plan}}$ must output whether the plan is correct or not $B_{\text{plan}} \in \{0, 1\}$:
\begin{equation}
    \text{VLM}_{\text{plan}}(I_{\text{start}}, T, P) \rightarrow B_{\text{plan}}
\end{equation}

For execution verification, given the task goal $T$, a subtask description $P_i$, and the visual observations before and after execution, $I_{\text{start}}$ and $I_{\text{end}}$, the model $\text{VLM}_{\text{exec}}$ similarly outputs $B_{\text{exec}} \in \{0, 1\}$:
\begin{equation}
    \text{VLM}_{\text{exec}}(I_{\text{start}}, I_{\text{end}}, T, P_i) \rightarrow B_{\text{exec}}
\end{equation}

\subsection{Model Architecture} 

The \emph{Guardian} model is built upon the open-source VLM  InternVL3-8B~\cite{zhu2025internvl3exploringadvancedtraining}. 
As shown in \cref{fig:model_architecture} (left), it comprises three components: a text tokenizer that converts text into discrete token embeddings, a visual encoder (InternViT-300M) that transforms individual images into visual embeddings, and a transformer-based LLM (Qwen2.5-7B) that processes the concatenated multimodal tokens to predict the answer.

Rather than concatenating multiple images into a single grid-based image as in prior work~\cite{duan2025aha,ifailsense2026}, \emph{Guardian} processes each image independently through the visual encoder. This design preserves fine-grained spatial details within each image and allows to explicitly reason about spatial and temporal changes for more accurate failure detection. Furthermore, instead of directly outputting classifications~\cite{duan2025aha,pmlr-v232-du23b,ifailsense2026}, \emph{Guardian} generates an explicit reasoning trace before concluding success or failure.

\subsection{Model Training} 
\label{sec:modeltraining}

We fine-tune \emph{Guardian} on \emph{GuardianFail-36k} using parameter-efficient Low-Rank Adaptation (LoRA)~\cite{hu2022lora}, while freezing the visual encoder. Training minimizes cross-entropy loss for next-token prediction.

Although CoT has shown promise to improve performance, it brings additional computation overhead.
Inspired by prior work~\cite{chen2025trainingstrategiesefficientembodied}, we explore three strategies for incorporating CoT into failure detection:
(1) \emph{Vanilla:} a baseline model trained and evaluated to directly predict final answers (\emph{A}) without \emph{CoT};
(2) \emph{Thinking:} the model is trained and inferred with explicit reasoning, always generating \emph{CoT} before \emph{A}; 
(3) \emph{Dropout:} in training, the model alternates between generating \emph{CoT}+\emph{A} and directly predicting \emph{A}, while at test time, only \emph{A} is produced. 
Results in~\Cref{tab:trainingstrategies} show that adding reasoning traces consistently improves performance. The Thinking strategy performs best but increases inference time, while Dropout offers a better speed-accuracy trade-off.

\subsection{Integration into Robotic Manipulation} 

\emph{Guardian} can be seamlessly plugged into existing robotic manipulation pipelines as a verification layer without requiring any architectural modification.
Without loss of generality, consider a modular robotic manipulation framework. As shown in \cref{fig:model_architecture} (right), \emph{Guardian} can be inserted at each planning and subtask execution step to detect potential failures. 
Upon detection, it triggers replanning or re-executes the corresponding motion policy to facilitate recovery, and uses its fine-grained failure reasoning as a hint to replan.
\section{Experiments}
\label{sec:expr}

\subsection{Experimental Setup}

\noindent \textbf{Evaluation datasets.}
Our main evaluation focuses on three unseen real-world benchmarks:
\emph{RoboFail}~\cite{liu2023reflect}, \emph{UR5-Fail}, and \emph{RoboVQA}~\cite{sermanet2024}. 
\emph{RoboFail} is a manually curated single-view UR5 failure dataset.
\emph{UR5-Fail} is our constructed multi-view real-robot dataset.
\emph{RoboVQA (RVQA)} is single-view and spans three embodiments: an Everyday Robot, a human arm, and a human using a grasping tool.
In ablations, we additionally report results on GuardianFail-36k testing splits. 
We use average classification accuracy as the metric.

\noindent \textbf{Implementation details.}
We fine-tune models using LoRA with AdamW, bf16 precision, and a cosine schedule. The hyperparameters are listed in the supplementary materials. 
Training is conducted on 4 H100 GPUs for 6 hours using GuardianFail-36k unless otherwise specified. 
For RLBench-Fail, we randomly sample one or four views during training to mitigate view-specific overfitting. 
The best checkpoint is selected via validation accuracy. Models use a zero temperature and greedy decoding to produce identical outputs at inference-time across runs.

\begin{table}[t]
\centering
\caption{Comparison of failure detection models on unseen real-world benchmarks. Execution and planning accuracies are reported. $^*$ denotes numbers from the original paper.}
\label{tab:comparison_in_ood}

\resizebox{\linewidth}{!}{
\large
\begin{tabular}{l c >{\columncolor[HTML]{e5eff8}}c >{\columncolor[HTML]{fbedde}}c >{\columncolor[HTML]{e5eff8}}c >{\columncolor[HTML]{fbedde}}c >{\columncolor[HTML]{e5eff8}}c}
\toprule
\textbf{Model} &
\textbf{Trained on} &
\multicolumn{2}{c}{\textbf{RoboFail~\cite{liu2023reflect}}} &
\multicolumn{2}{c}{\textbf{UR5-Fail}} &
\multicolumn{1}{c}{\textbf{RVQA~\cite{sermanet2024}}} \\
\cmidrule(lr){3-4} 
\cmidrule(lr){5-6} 
\cmidrule(lr){7-7}
& \textbf{GuardianFail} 
& \textbf{Exec} 
& \textbf{Plan} 
& \textbf{Exec} 
& \textbf{Plan} 
& \textbf{Exec} \\
\midrule

\rowcolor{gray!15}
\multicolumn{7}{l}{\textit{Closed-Source VLM}} \\

GPT-4o
& \textcolor{red}{\ding{55}}
& 0.80 & 0.67 & \textbf{0.77} & 0.85 & 0.79 \\

GPT-4o {\normalsize +Sentinel-Video-QA~\cite{agia2024unpackingfailuremodesgenerative}}
& \textcolor{red}{\ding{55}}
& 0.80 & 0.63 & 0.76 & 0.62 & 0.66 \\

\midrule

\rowcolor{gray!15}
\multicolumn{7}{l}{\textit{Robotic Failure Detection VLMs}} \\

RoboFAC-7B~\cite{robofac2025}
& \textcolor{red}{\ding{55}}
& 0.25 & 0.05 & 0.54 & 0.02 & 0.52 \\

AHA-13B$^*$~\cite{duan2025aha}
& \textcolor{red}{\ding{55}}
& 0.64* & - & - & - & - \\

I-Fail-Sense-3B~\cite{ifailsense2026}
& \textcolor{red}{\ding{55}}
& 0.43 & 0.67 & 0.47 & 0.46 & 0.53 \\

Cosmos-Reason2-8B~\cite{nvidia2026cosmosreason2}
& \textcolor{red}{\ding{55}}
& 0.78 & 0.53 & 0.59 & 0.67 & 0.76 \\
\midrule

CLIP+MLP~\cite{pmlr-v139-radford21a}
& \textcolor{green}{\ding{51}}
& 0.42 & 0.43 & 0.51 & 0.51 & 0.52 \\

I-Fail-Sense-3B~\cite{ifailsense2026}
& \textcolor{green}{\ding{51}}
& 0.76 & 0.52 & 0.55 & 0.6 & 0.58 \\

Cosmos-Reason2-8B~\cite{nvidia2026cosmosreason2}
& \textcolor{green}{\ding{51}}
& 0.82 & \textbf{0.70} & 0.65 & 0.83 & 0.77 \\

\midrule
Guardian-8B 
& \textcolor{green}{\ding{51}}
& \textbf{0.86} & \textbf{0.70} & \textbf{0.77} & \textbf{0.89} & \textbf{0.85} \\

\bottomrule
\end{tabular}
}
\label{tab:ood_domain_eval}

\end{table}

\subsection{Comparison with State of the Art}
\label{sec:sotacomparison}


\noindent \textbf{Compared methods.}
We compare against GPT-4o~\cite{openai2024gpt4ocard} and specialized robotic failure detectors including
Cosmos-Reason2-8B~\cite{nvidia2026cosmosreason2}, 
AHA-13B~\cite{duan2025aha}, 
RoboFAC-7B~\cite{robofac2025}, 
I-Fail-Sense-3B~\cite{ifailsense2026}, 
Sentinel-Video-QA~\cite{agia2024unpackingfailuremodesgenerative}, 
and CLIP+MLP~\cite{pmlr-v139-radford21a}. 
AHA results come from the original paper, as the model is not publicly released.
For the other methods, we run their released checkpoints or train models with their codebase.

\noindent \textbf{Results.}
Table~\ref{tab:comparison_in_ood} reports performance on the test sets. GPT-4o achieves strong performance due to its scale and general reasoning ability.
However, applying the Sentinel-Video-QA~\cite{agia2024unpackingfailuremodesgenerative} self-interrogation prompting degrades accuracy as it constrains the reasoning ability of the original model. 
Models trained exclusively on simulated failures (RoboFAC~\cite{robofac2025} and AHA~\cite{duan2025aha}) show limited transfer to real-robot benchmarks. 
Both rely on simulation-only perturbations, which likely restrict generalization to unseen real-world manipulators and sensor noise. 
I-Fail-Sense~\cite{ifailsense2026} is trained on both simulation and real-world trajectories, but its supervision focuses primarily on semantic misalignment detection rather than structured planning or low-level control failures, which likely limits its performance on the benchmarks. 
Cosmos-Reason~\cite{nvidia2026cosmosreason2} performs competitively, reflecting strong physical reasoning capabilities, but it is optimized for broad embodied reasoning rather than only for failure verification.

Since prior models are trained on different datasets, we further fine-tune representative open-source models on the same GuardianFail-36k dataset to isolate the effects of data and architecture (Table~\ref{tab:comparison_in_ood}). We also include a lightweight CLIP+MLP baseline, which performs substantially worse, highlighting the necessity of large vision–language models. 
Notably, training on GuardianFail-36k consistently improves all methods,  underscoring the importance of well-curated, cross-environment data with broad failure coverage.

Guardian achieves the strongest overall performance across RoboFail, UR5-Fail, and RoboVQA. 
Unlike I-Fail-Sense~\cite{ifailsense2026}, Guardian preserves multi-view spatial structure and produces explicit chain-of-thought reasoning, enabling structured subtask-level verification. 
Compared to Cosmos-Reason, which is pretrained for embodied reasoning and primarily developed in single-view settings, Guardian leverages an InternVL backbone with explicit multi-view supervision, which likely explains its stronger fine-grained failure detection performance even under identical training data. 

An analysis of Guardian's detection confusion matrices and failure cases is provided in the supplementary materials.

\begin{table}[t]
\vspace{0.7em}
\centering
\small
\caption{Impact of the Guardian fine-tuning data mix on the binary accuracy averaged over planning and execution.}
\label{tab:ablation_training_mixture}
\setlength{\tabcolsep}{4pt}
\begin{adjustbox}{max width=0.9\linewidth}
\begin{tabular}{cc|ccccc}
\toprule
\multicolumn{2}{c|}{\textbf{Training Data}} & 
\textbf{RLBench} & \textbf{BDV2} & \textbf{Robo} & \textbf{UR5} & \textbf{Robo} \\
\cmidrule(lr){1-2}
\textbf{RLBench} & \textbf{BDV2} & 
\textbf{-Fail} & \textbf{-Fail} & \textbf{-Fail} & \textbf{-Fail} & \textbf{-VQA} \\
\midrule
\textcolor{red}{\ding{55}} & \textcolor{red}{\ding{55}} & 
0.65 & 0.69 & 0.65 & 0.73 & 0.75 \\
\textcolor{green}{\ding{51}} & \textcolor{red}{\ding{55}} & 
0.82 & 0.70 & 0.69 & 0.72 & 0.66 \\
\textcolor{red}{\ding{55}} & \textcolor{green}{\ding{51}} & 
0.65 & 0.86 & 0.71 & 0.68 & 0.77 \\
\textcolor{green}{\ding{51}} & \textcolor{green}{\ding{51}} & 
\textbf{0.85} & \textbf{0.88} & \textbf{0.78} & \textbf{0.83} & \textbf{0.85} \\
\bottomrule
\end{tabular}
\end{adjustbox}
\end{table}
\begin{figure}[t]
\centering
\includegraphics[width=0.6\linewidth]{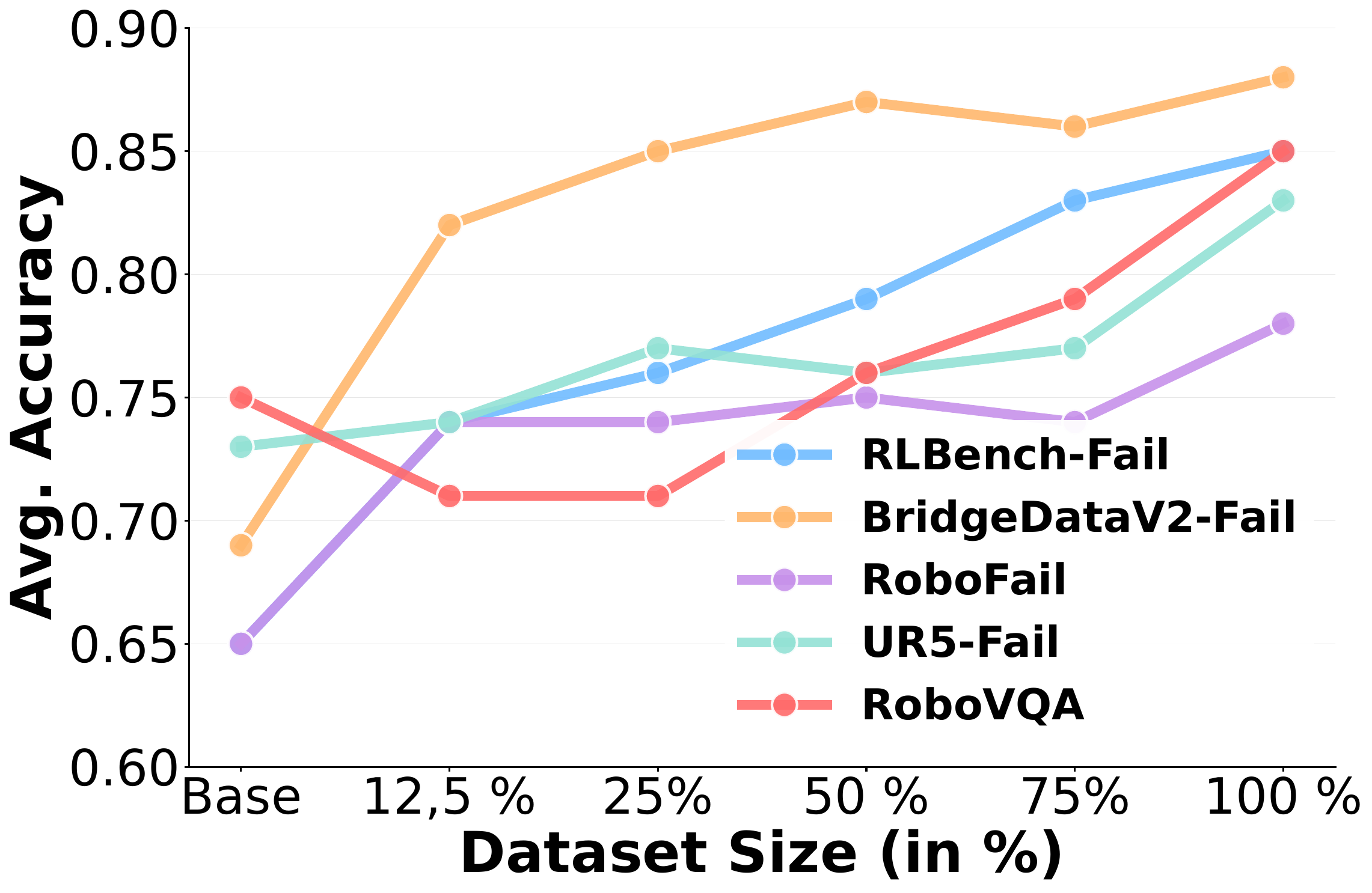}
\caption{GuardianFail training data size impact on the binary accuracy averaged over planning and execution.
}
\label{fig:scaling_law}
\vspace{-0.5em}
\end{figure}

\subsection{Failure Data Ablations}
\label{sec:data-ablations}

We next analyze how training data composition (simulation and real data), and dataset scale influence cross-environment generalization while keeping the architecture fixed.

\noindent \textbf{Data composition.}
Table~\ref{tab:ablation_training_mixture} compares InternVL3-8B without fine-tuning, with single-source training, and with the full GuardianFail-36k dataset. Without fine-tuning, performance is moderate. Training only on RLBench-Fail improves simulated results but transfers weakly to real-robot datasets. Training only on BridgeDataV2-Fail (BDV2) improves real-world performance but shows limited simulation transfer. Combining both datasets yields consistent gains across RoboFail, UR5-Fail, and RoboVQA. The same trend holds for other architectures in Table~\ref{tab:comparison_in_ood}, emphasizing the importance of cross-environment composition and broad coverage of failure supervision.

\noindent \textbf{Dataset scaling.}
Fig.~\ref{fig:scaling_law} demonstrates consistent scaling behavior as the amount of GuardianFail-36k data increases. Performance on in-domain and unseen real-world datasets improves steadily with dataset size. This indicates that scaling structured failure generation remains a promising direction for improving cross-environment generalization.
\begin{table}[t]
\centering
\caption{Comparison of training and test-time strategies with and without CoT (\cref{sec:modeltraining}). ``A'' denotes the final answer. We report binary accuracy and inference time (seconds/sample) on one H100, averaged across all test sets.}
\label{tab:trainingstrategies}
\vspace{-0.5em}
\setlength{\tabcolsep}{10pt}
\renewcommand{\arraystretch}{0.9}
\resizebox{\linewidth}{!}{%
\begin{tabular}{@{}lccccc@{}}
\toprule
& \multicolumn{2}{c}{\textbf{Output}} 
& \multicolumn{2}{c}{\textbf{Avg. Accuracy}} 
& \textbf{Inf.} \\
\cmidrule(lr){2-3} \cmidrule(lr){4-5} \cmidrule(l){6-6}
\textbf{Model} & \textbf{Train} & \textbf{Test} & \textbf{Exec} & \textbf{Plan} & \textbf{Time} \\
\midrule
Vanilla  & A & A & 0.80 & 0.78 & 0.68s \\
Dropout  & \textbf{CoT}, A $\vee$ A & A & 0.81 & 0.83 & 0.68s \\
Thinking & \textbf{CoT}, A & \textbf{CoT}, A & \textbf{0.83} & \textbf{0.84} & 4.3s \\
\bottomrule
\end{tabular}%
}
\end{table}

\subsection{Failure Detection Method Ablations}
\label{sec:method-ablations}

We further isolate the contribution of architectural and reasoning design choices on failure detection performance.

\noindent \textbf{Train-time and test-time CoT strategies.}
Table~\ref{tab:trainingstrategies} compares Vanilla, Dropout, and Thinking strategies for integrating CoT reasoning traces.
Training with reasoning traces consistently improves accuracy for both execution and planning.
The Thinking strategy achieves the highest accuracy but increases inference time by approximately $6\times$.
Dropout offers a favorable trade-off, retaining most of the accuracy gains without additional test-time cost.

\begin{table}[t]
\centering
\begin{minipage}[c]{0.2\textwidth}
\caption{Impact of image representation. We report accuracy averaged across all datasets, for both Execution and Planning. AHA~\cite{duan2025aha} and I-Fail-Sense~\cite{ifailsense2026} use multi-view concatenation.}
\label{tab:ablation_image_format}
\end{minipage}
\hspace{0.01\textwidth}
\begin{minipage}[c]{0.24\textwidth}
\centering
\setlength{\tabcolsep}{3.5pt}
\renewcommand{\arraystretch}{0.95}
\footnotesize
\begin{tabular}{cccc}
\toprule
\textbf{Views} & \textbf{Image Format} & \textbf{Exec} & \textbf{Plan} \\
\midrule
\multirow{2}{*}{Single}
& concat    & 0.69 & 0.82 \\
& separated & 0.72 & 0.82 \\
\midrule
\multirow{2}{*}{Multi}
& concat    & 0.74 & 0.82 \\
& separated & \textbf{0.83} & \textbf{0.84} \\
\bottomrule
\end{tabular}
\end{minipage}
\vspace{-1em}
\end{table}
\begin{figure}[t]
\centering
\includegraphics[width=\linewidth]{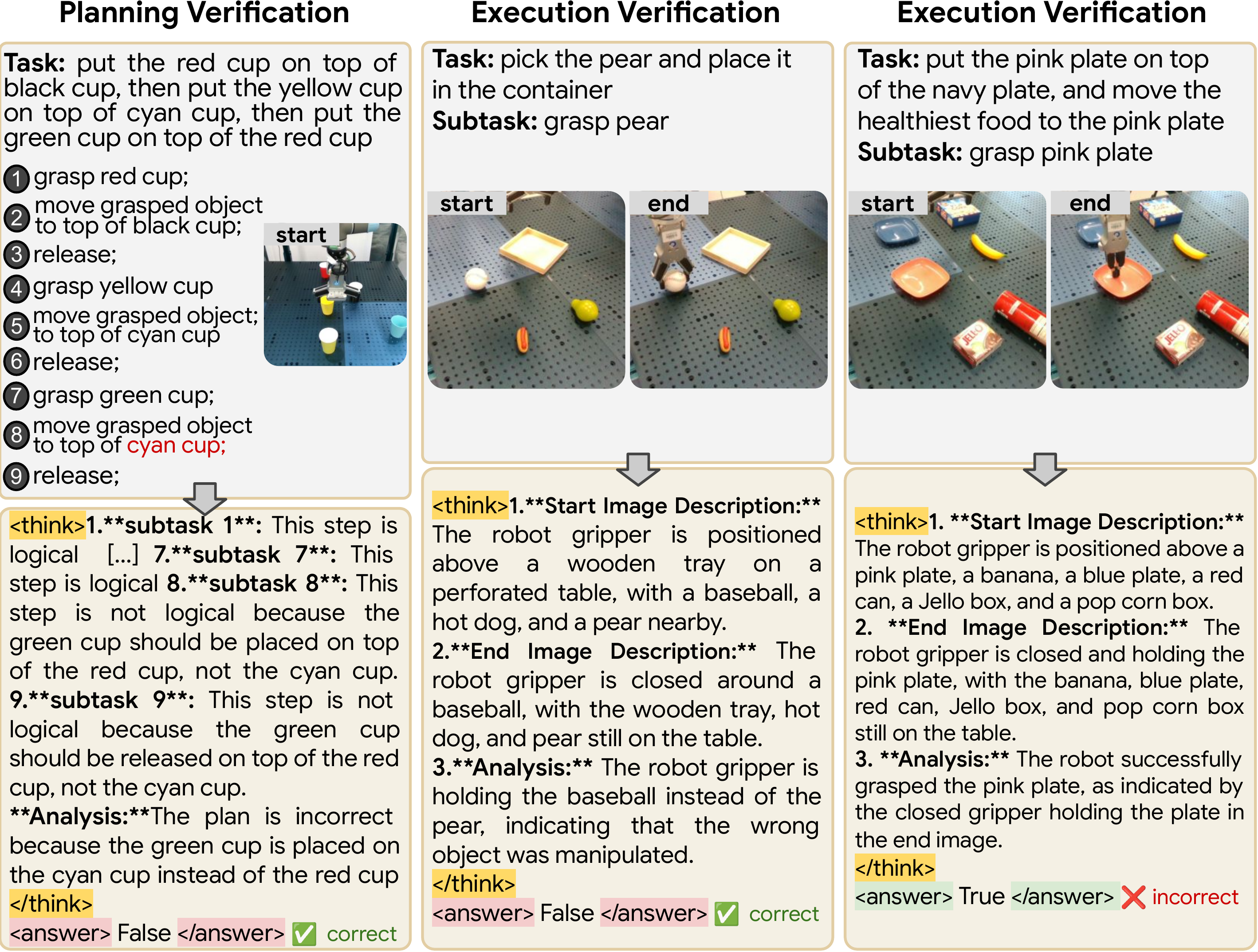}
\caption{Qualitative examples of Guardian on UR5-Fail. The left and middle examples show successful plan and execution verification, while the right example shows a failure case.
}
\label{fig:UR5Fail_qualitative_examples}
\vspace{-1em}
\end{figure}
\noindent \textbf{Image representation.}
Prior methods (AHA~\cite{duan2025aha}, I-Fail-Sense~\cite{ifailsense2026}) concatenate multiple views into a single grid image. Therefore, \cref{tab:ablation_image_format} isolates the effect of the visual input representation by varying the number and encoding of views. Processing multi-views as separate high-resolution images consistently outperforms concatenated grid representations, which leads to visual clutter.
With multi-views, separated inputs improve execution accuracy from 0.74 (I-Fail-Sense and AHA concatenation) to 0.83, while planning accuracy increases more modestly from 0.82 to 0.84.
Even in single-view, separating start and end images yields a small execution gain (0.69 to 0.72).
The smaller planning gain is expected since planning samples contain only the front view.

\subsection{Qualitative examples}
\label{sec:qualitative-examples}

\cref{fig:UR5Fail_qualitative_examples} shows qualitative results of \emph{Guardian} on \emph{UR5-Fail}. The left (planning) and middle (execution) examples illustrate correct detection of an invalid plan and a wrong-object manipulation. The example on the right presents a failure case where Guardian misjudges the grasp as successful, despite the robot failing to grasp the plate.

\begin{figure*}[ht]
\centering
\small
\captionof{table}{Success rate of 3D-LOTUS++ across unseen RLBench tasks without and with different VLM-based failure detectors. The last row denotes our full Guardian model.}
\setlength{\tabcolsep}{3pt}
\resizebox{\textwidth}{!}{
\begin{tabular}{@{}c c c c c c c c c c c c c c@{}}
\toprule
Verifier
& \makecell[c]{Trained on \\ GuardianFail}
& \makecell[c]{CoT}
& \makecell[c]{Average}
& \makecell[c]{Open \\ top drawer}
& \makecell[c]{Push \\ white button}
& \makecell[c]{Push \\ 4 buttons}
& \makecell[c]{Lift \\ red duck}
& \makecell[c]{Screw \\ maroon bulb}
& \makecell[c]{Slide block \\ to yellow target}
& \makecell[c]{Lift \\ black block}
& \makecell[c]{Close \\ grill}
& \makecell[c]{Close \\ microwave}
& \makecell[c]{Close \\ bottom drawer} \\
\midrule
\textcolor{red}{\ding{55}}
& \textcolor{red}{\ding{55}}
& \textcolor{red}{\ding{55}}
& $0.45_{\pm 0.03}$
& $0.54_{\pm 0.03}$
& $0.92_{\pm 0.01}$
& $0.09_{\pm 0.03}$
& $0.38_{\pm 0.05}$
& $0.50_{\pm 0.04}$
& $0.07_{\pm 0.03}$
& $0.85_{\pm 0.02}$
& $0.11_{\pm 0.02}$
& $0.97_{\pm 0.01}$
& $0.06_{\pm 0.02}$ \\
\textcolor{green}{\ding{51}}
& \textcolor{red}{\ding{55}}
& \textcolor{red}{\ding{55}}
& $0.49_{\pm 0.02}$
& $0.64_{\pm 0.02}$
& $0.93_{\pm 0.02}$
& $0.12_{\pm 0.03}$
& $0.40_{\pm 0.04}$
& $0.48_{\pm 0.01}$
& $0.15_{\pm 0.03}$
& $0.86_{\pm 0.03}$
& $0.12_{\pm 0.02}$
& $0.98_{\pm 0.00}$
& $0.19_{\pm 0.04}$ \\
\textcolor{green}{\ding{51}}
& \textcolor{green}{\ding{51}}
& \textcolor{red}{\ding{55}}
& $0.51_{\pm 0.04}$
& $0.70_{\pm 0.06}$
& $0.92_{\pm 0.01}$
& $\mathbf{0.18}_{\pm 0.03}$
& $0.40_{\pm 0.03}$
& $0.48_{\pm 0.08}$
& $0.20_{\pm 0.04}$
& $0.86_{\pm 0.03}$
& $0.16_{\pm 0.04}$
& $\mathbf{1.00}_{\pm 0.00}$
& $0.19_{\pm 0.05}$ \\
\midrule
\makecell[c]{\textcolor{green}{\ding{51}}}
& \textcolor{green}{\ding{51}}
& \textcolor{green}{\ding{51}}
& $\mathbf{0.54}_{\pm 0.03}$
& $\mathbf{0.75}_{\pm 0.02}$
& $\mathbf{0.96}_{\pm 0.02}$
& $\mathbf{0.18}_{\pm 0.03}$
& $\mathbf{0.42}_{\pm 0.03}$
& $\mathbf{0.54}_{\pm 0.04}$
& $\mathbf{0.31}_{\pm 0.06}$
& $\mathbf{0.89}_{\pm 0.04}$
& $\mathbf{0.18}_{\pm 0.03}$
& $\mathbf{1.00}_{\pm 0.00}$
& $\mathbf{0.20}_{\pm 0.04}$ \\
\bottomrule
\vspace{0.1em}
\end{tabular}
}
\label{tab:sim_failure_detectors_selected_tasks}

\includegraphics[width=0.95\linewidth]{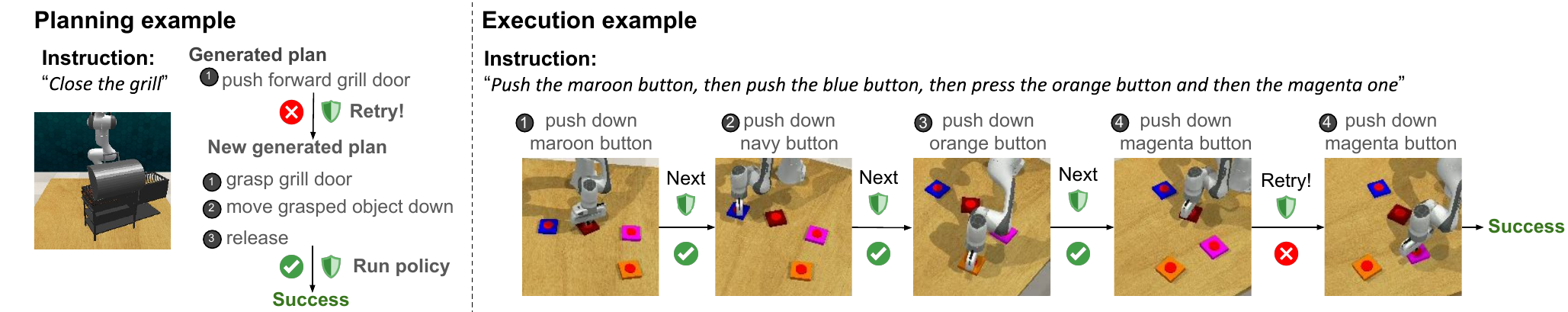}
\captionof{figure}{Verification with Guardian during online task execution on RLBench. The failure reasoning is fed back into the system to support replanning. Left: Successful refinement of an incorrect plan. Right: Successful correction of a subtask execution.}
\label{fig:rlbench_online_examples}
\vspace{-1.2em}
\end{figure*}
\subsection{Downstream Robotic Tasks}

Finally, we evaluate whether improved failure detection translates into tangible gains in online robot experiments. We integrate Guardian as a verification module, as illustrated in \cref{fig:model_architecture} (right), into 3D-LOTUS++ on unseen tasks, both in simulation and in the real world. We compare the base manipulation policy performance without and with a verifier. 
As a verifier, we use InternVL3-8B, either off-the-shelf without robot failure fine-tuning, or fine-tuned on GuardianFail-36k datasets with and without reasoning. 
Guardian corresponds to the setting where the verifier InternVL3-8B is fine-tuned on GuardianFail-36k and leverages structured step-by-step reasoning. 
At each planning and subtask stage, Guardian verifies correctness; upon detecting failure, we inject back the failure reasoning trace to help with replanning, and we retry up to three times.

\noindent \textbf{Simulation:} We evaluate on 10 tasks on RLBench as shown in \cref{fig:rlbench_online_examples}, selecting unseen tasks not used for GuardianFail-36k training, spanning various motion primitives and objects. For each task, we run 100 episodes (20 per seed across 5 seeds) and report the mean success rate ($\pm$ standard deviation). As shown in Table~\ref{tab:sim_failure_detectors_selected_tasks}, success rates increase consistently as stronger verification models are used. Guardian provides the largest gains, improving the average success rate from $0.45$ (no verifier) to $0.54$. Improvements are particularly pronounced on long-horizon or error-prone tasks, indicating that structured failure reasoning provides reliable recovery signals during online execution.

\begin{table}[t]
\centering
\small
\caption{Success rate of 3D-LOTUS++ on real robot experiments without and with different VLM-based failure detectors. The last row denotes our full Guardian model.}
\setlength{\tabcolsep}{3pt}
\resizebox{\linewidth}{!}{
\begin{tabular}{@{}c c c cc cc cc@{}}
\toprule
\multirow{2}{*}{Verifier}
& \multirow{2}{*}{\makecell[c]{Trained on \\ GuardianFail}}
& \multirow{2}{*}{CoT}
& \multicolumn{2}{c}{Put food}
& \multicolumn{2}{c}{Arrange fruits}
& \multicolumn{2}{c}{Stack cups} \\
\cmidrule(lr){4-5} \cmidrule(lr){6-7} \cmidrule(lr){8-9}
& &
& Norm & Pert
& Norm & Pert
& Norm & Pert \\
\midrule

\textcolor{red}{\ding{55}}
& \textcolor{red}{\ding{55}}
& \textcolor{red}{\ding{55}}
& 15/20 & 4/20
& 10/20 & 3/20
& 9/20 & 2/20 \\

\textcolor{green}{\ding{51}}
& \textcolor{red}{\ding{55}}
& \textcolor{red}{\ding{55}}
& 12/20 & 10/20
& 9/20 & 9/20
& 7/20 & 6/20 \\

\textcolor{green}{\ding{51}}
& \textcolor{green}{\ding{51}}
& \textcolor{red}{\ding{55}}
& 17/20 & 13/20
& \textbf{14/20} & \textbf{12/20}
& 10/20 & 8/20 \\

\midrule
\makecell[c]{\textcolor{green}{\ding{51}} }
& \textcolor{green}{\ding{51}}
& \textcolor{green}{\ding{51}}
& \textbf{18/20} & \textbf{15/20}
& \textbf{14/20} & \textbf{12/20}
& \textbf{12/20} & \textbf{11/20} \\

\bottomrule
\end{tabular}
}
\label{tab:realrobot_online_exp}
\vspace{-1em}
\end{table}
\noindent \textbf{Real robot:}
We deploy Guardian zero-shot on a 6-DoF UR5 with three RealSense D435 cameras. We evaluate three unseen tasks: placing food in a box, arranging fruits on colored plates, and stacking colored cups. 
For each task, we run 40 episodes (20 under normal conditions and 20 with perturbations). 
Perturbations are human-induced during execution, involving translating objects or swapping an object with the wrong one.
Table~\ref{tab:realrobot_online_exp} shows consistent improvements in both nominal and perturbed settings. 
Under perturbations, gains are especially large, demonstrating that structured failure supervision enhances robustness under distribution shifts and execution noise. 
Without dedicated robot failure data fine-tuning, InternVL3-8B performs poor distinction between success and failure, which damages the base robot policy performance in nominal cases. 
We observe that training on GuardianFail-36k significantly improves the performance, suggesting that structured failure reasoning traces augment the verifier's spatial understanding. Please check the supplementary video for real-robot visualizations.

\section{Conclusion}
\label{sec:discussion}

This work tackles the scarcity of data for training robotic failure detection models. 
We introduce an automated method that generates diverse planning and execution failures with interpretable reasoning annotations in both simulation and real-world environments. This enables the construction of GuardianFail-36k with over 36K examples, which substantially expands the diversity and coverage of existing failure datasets. We train Guardian, a failure detection VLM fine-tuned on GuardianFail-36k, leveraging multi-view image inputs and step-by-step reasoning. Guardian achieves state-of-the-art generalization performance on three real-world benchmarks, including RoboFail, RoboVQA, and our constructed UR5-Fail. 
We further demonstrate Guardian’s zero-shot plug-and-play utility on unseen tasks by improving an LLM-based manipulation system in both simulation and real-robot deployments. Future work will investigate incorporating rich failure feedback directly into policy learning to enable more robust recovery.

\section{Acknowledgments}
\label{sec:acknowledgments}

This work was performed using HPC resources from GENCI-IDRIS (Grant 2025-AD011015795[R1] and AD011018232). It was funded in part by the French government under management of Agence Nationale de la Recherche as part of the “France 2030" program, reference ANR-23-IACL-0008 (PR[AI]RIE-PSAI project), the ANR project 3D-GEM (ANR-25-CE23-7777-01), the ANR project VideoPredict ANR-21-FAI1-0002- 01. Cordelia Schmid would like to acknowledge the support by the Körber European Science Prize.

{
    \small
    \bibliographystyle{ieeenat_fullname}
    \bibliography{main}
}

\clearpage
\appendix
\section*{Supplementary Material}


First, Section \ref{sec:supp-mat-qualitative-examples} provides qualitative examples on the test sets, containing success and failure cases of the model. Second, Section \ref{sec:supp-mat-failure-cases-analysis} analyzes the failure cases of Guardian quantitatively. Then, we analyze the quality of the predicted chain-of-thought in \ref{sec:supp-mat-cot-quality}. Next, Section \ref{sec:supp-mat-real-robot} shows the real-robot setup for the real-world experiments. Finally, Section \ref{sec:supp-mat-implementation-details} lists implementation details for reproducibility.


\begin{figure}[b]
\centering
\includegraphics[width=\linewidth]{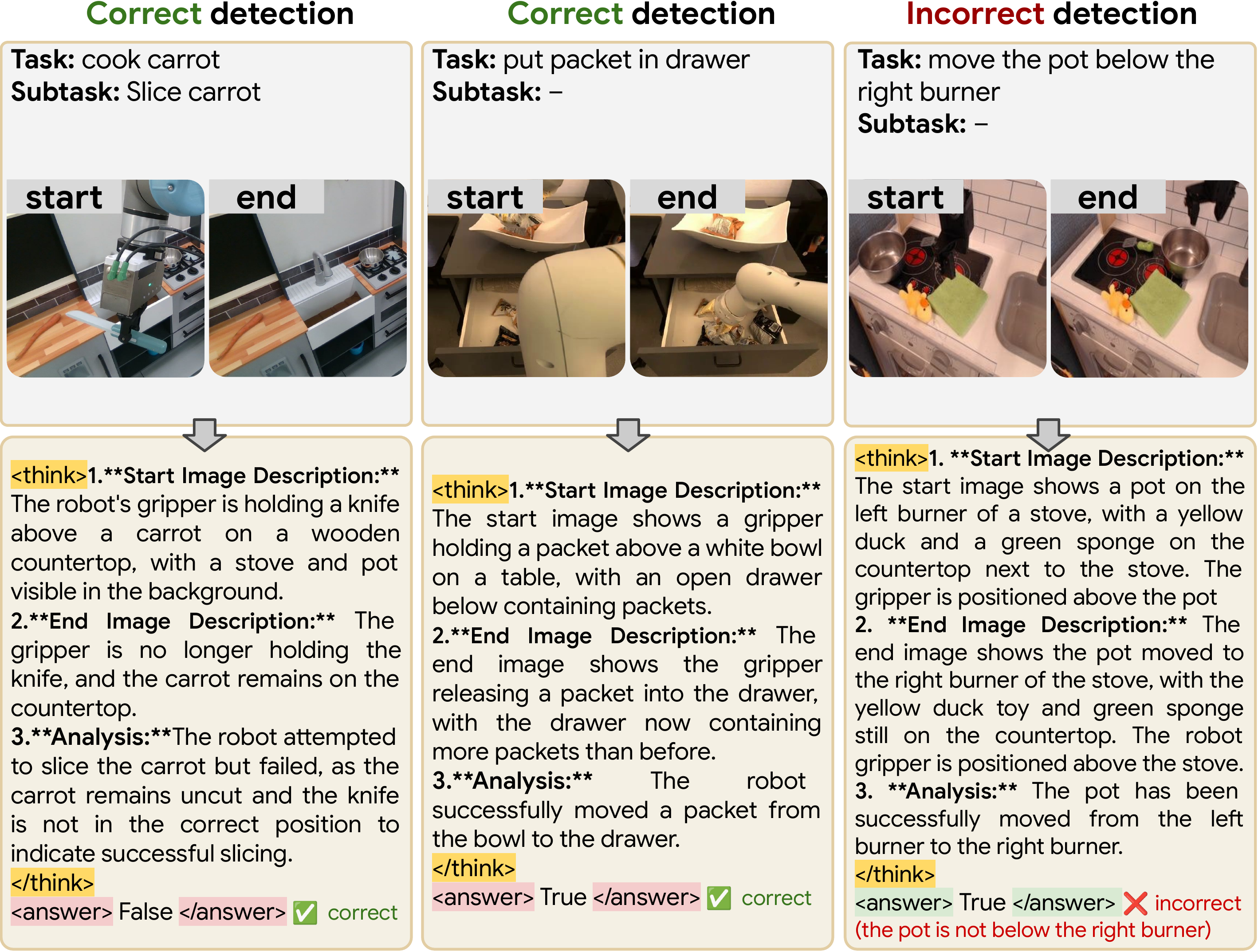}
\caption{Execution Verification. Qualitative examples of Guardian on RoboFail~\cite{liu2023reflect} (left), RoboVQA~\cite{sermanet2024} (middle), and BridgeDataV2-Fail (right). The left and middle are correct detection examples, the right is a failure case of the model.
}
\label{fig:execution_qualitative_examples_robofail_robovqa_bdv2fail}
\vspace{-1em}
\end{figure}
\begin{figure}[t]
\centering
\includegraphics[width=\linewidth]{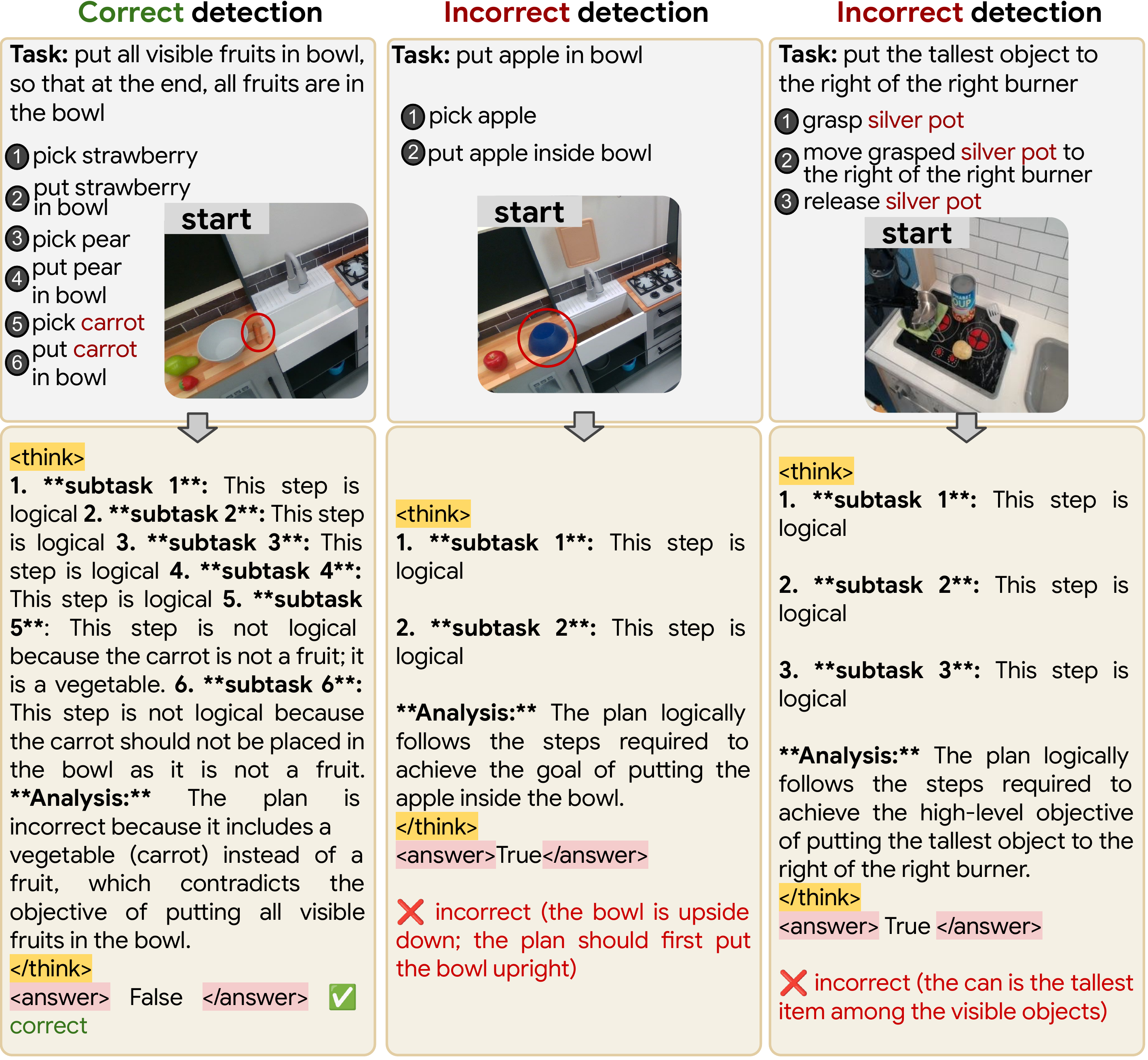}
\caption{Planning Verification. Qualitative examples of Guardian on RoboFail~\cite{liu2023reflect} (left and middle), BridgeDataV2-Fail (right). The left is a correct detection example, the middle and right are failure cases of the model.
}
\label{fig:planning_qualitative_examples_robofail_robovqa_bdv2fail}
\vspace{-1em}
\end{figure}
\section{Qualitative Examples and Failure Cases}
\label{sec:supp-mat-qualitative-examples}

We provide qualitative examples of Guardian's execution verification in \cref{fig:execution_qualitative_examples_robofail_robovqa_bdv2fail} and planning verification in \cref{fig:planning_qualitative_examples_robofail_robovqa_bdv2fail}. The visualizations contain both correct detection cases and incorrect ones to show failure cases of the model. 

In the execution verification figure, the right example is a failure case: the model correctly understands the pot has been moved to the right burner, but does not detect that the pot is above and not below the burner.

In the planning verification figure, the middle and right examples are failure cases, in which either a subtask is missing (middle example) or a wrong object is manipulated (right example). In both cases, again, the spatial reasoning of the model is failing to detect that the object is upside-down or to compare the heights of the items.

This motivates future work to incorporating more spatial reasoning cases in the training data or co-training on large-scale embodied VQA data. We further provide a quantitative analysis of the failure cases in Section \ref{sec:supp-mat-failure-cases-analysis}.

\begin{figure}[t]
    \vspace{0.4em}
    \centering
    \includegraphics[width=\linewidth]{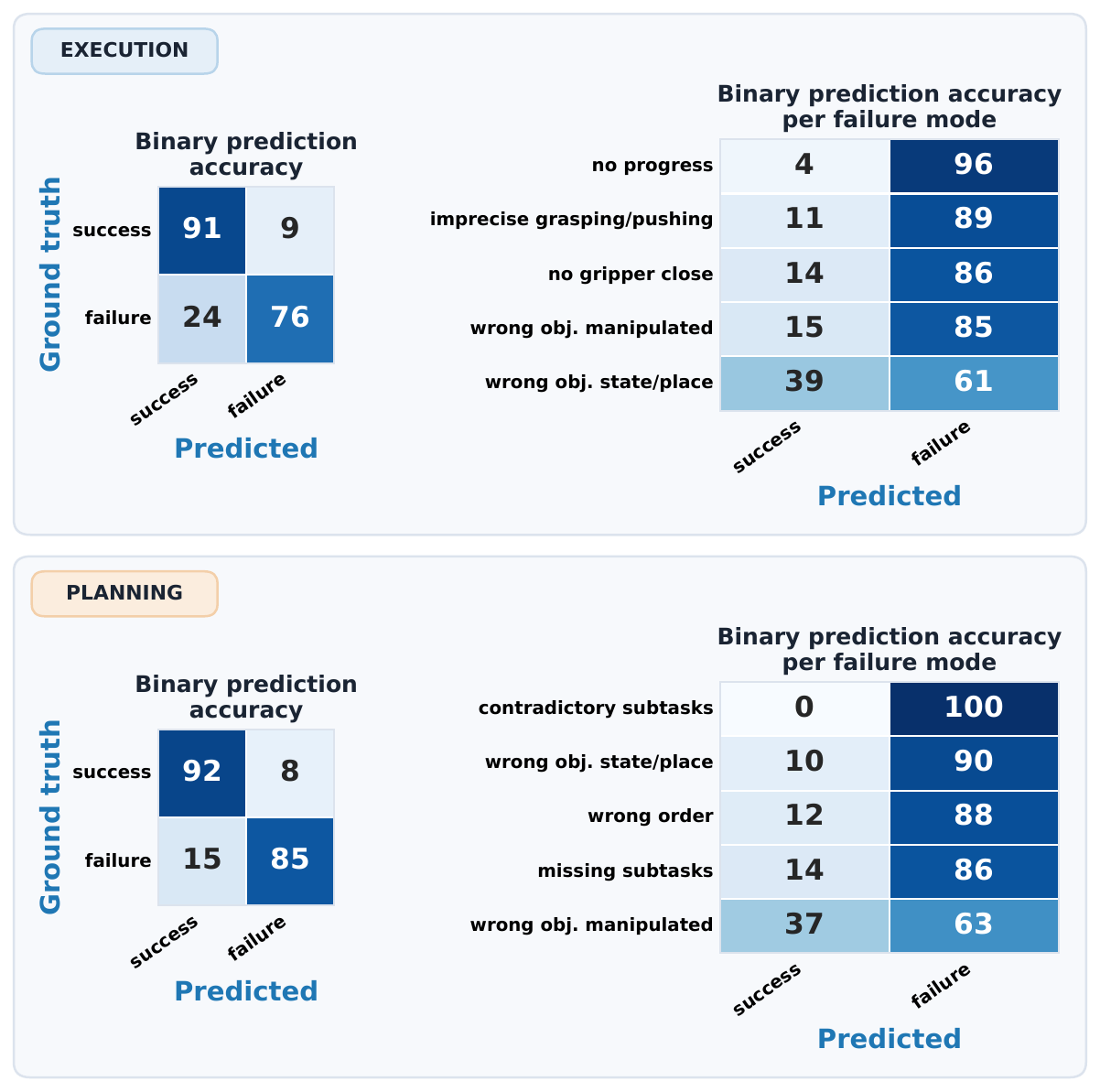}
    \vspace{-1.9em}
    \caption{Confusion matrices for Guardian's failure detection on execution (top) and planning (bottom). For each scenario, the left matrix reports binary success/failure prediction accuracy, and the right matrix breaks this accuracy down per ground-truth failure mode. Values are percentages; rows denote the ground-truth success or failure mode, and columns the binary class predicted.} 
    \label{fig:supp-mat-confusion_combined_binary}
\end{figure}

\section{Quantitative Analysis of the Failure Cases}
\label{sec:supp-mat-failure-cases-analysis}

Here, we analyze the cases where Guardian incorrectly classifies successes/failures in the test sets. We run the analysis on RLBench-Fail, BridgeDataV2-Fail, UR5-Fail, and RoboFail as they contain fine-grained failure categories, but RoboVQA does not. We show the confusion matrices of the model in \cref{fig:supp-mat-confusion_combined_binary}. 

\paragraph{Execution failures.}
Guardian correctly identifies $91\%$ of successful executions and $76\%$ of
failed ones (\cref{fig:supp-mat-confusion_combined_binary}, top), with a low
$9\%$ false-positive rate that avoids over-flagging nominal behavior.
Failures with a clear visual signature are detected almost perfectly: \textit{no
progress} ($96\%$), \textit{imprecise grasping/pushing} ($89\%$), \textit{no
gripper close} ($86\%$), \textit{wrong object manipulated} ($85\%$). Missed failures are concentrated in
\textit{wrong object state or placement} ($61\%$): here the correct object is
acted upon, and the action completes, but a wrong final pose betrays the failure, which requires more spatial reasoning to catch.

\textbf{Planning failures.}
The same pattern holds for planning
(\cref{fig:supp-mat-confusion_combined_binary}, bottom): $92\%$ accuracy on
valid plans and $85\%$ on invalid ones. Structurally inconsistent plans are
caught reliably: \textit{contradictory subtasks} ($100\%$), \textit{wrong order} ($88\%$),
\textit{missing subtasks} ($86\%$). The hardest case is again semantic where grounding the referent against the instruction reveals the error: even though $90\%$ of \textit{wrong
object state or placement} are correctly classified, the model has difficulty spotting
\textit{wrong object manipulated} ($63\%$) in subtasks.

\textbf{Takeaways.}
Guardian's errors are dominated by semantic mistakes --- wrong object
state/placement in execution, wrong object manipulated in planning --- whereas
physically or structurally obvious failures are flagged more reliably. The model is also more accurate at confirming successes ($91$--$92\%$)
than at flagging failures ($76$--$85\%$). One path to improvement could be increasing the emphasis on task-relevant object identification and goal state in the model's reasoning, together with stronger spatial reasoning via bounding boxes or segmentation

\section{Chain-of-Thoughts Quality Analysis}
\label{sec:supp-mat-cot-quality}

A natural question is whether Guardian's reasoning traces are reliable, and what they reveal about why the model fails. Such analysis leverages the explicit CoTs of the model, which would not be possible with a single binary output. 

\textbf{Setup.} To assess whether Guardian's reasoning traces are themselves reliable, we conduct a human study on 300
randomly drawn samples (100 each from UR5-Fail, RoboFail, and RoboVQA). For each
sample, an annotator is shown the multi-view images, task/subtask, full trace, and
ground-truth label, and answers two questions: (i) is the final predicted \texttt{<answer>}
consistent with the conclusion stated in the \textit{Analysis} step of the reasoning trace
(\textit{faithfulness}); and (ii) for misclassified samples, does the model's error first
arise in \textit{perception} (an inaccurate start/end image description) or in \textit{reasoning} (taking the descriptions as written, even if wrong, does the \textit{Analysis} reach the correct success/failure conclusion?). 

\textbf{Results.} We present results in \cref{tab:cot_faithfulness_and_error}. First, we look at the faithfulness of the traces. The traces are almost perfectly faithful: the final answer matches the analysis of the thinking part in $100\%$ of
cases on UR5-Fail, RoboFail, and RoboVQA, respectively. Guardian thus does not
contradict its own reasoning, so a trace can be read as a faithful account of how a
decision was reached. Next, we localize each misclassification to its first failing stage. Errors are dominated by perception rather than reasoning:
perception accounts for $75\%$, $68\%$, and $93\%$ of errors on UR5-Fail, RoboFail,
and RoboVQA, with the remainder attributable to reasoning over otherwise-correct
descriptions. In other words, when Guardian fails it is usually because it mis-describes visual details, rather than because it reasons incorrectly about what it saw, consistent with the spatial-grounding
errors illustrated in Section \ref{sec:supp-mat-qualitative-examples}.

\begin{table}[t]
\centering
\caption{Study of Guardian's reasoning traces. \textit{Faithfulness} is the rate at which
the final answer matches the conclusion of the \textit{thinking} step.
\textit{Perception} and \textit{Reasoning} report the share of
misclassifications whose first failing stage is an inaccurate image
description or an incorrect analysis of an otherwise-correct description.}
\label{tab:cot_faithfulness_and_error}
\setlength{\tabcolsep}{4pt}
\resizebox{\linewidth}{!}{%
\begin{tabular}{@{}lccc@{}}
\toprule
 & Faithfulness & \multicolumn{2}{c}{Error Attribution} \\
\cmidrule(lr){3-4}
Benchmark & ($\%$) & Perception ($\%$) & Reasoning ($\%$) \\
\midrule
UR5-Fail   & 100 & 75 & 25  \\
RoboFail~\cite{liu2023reflect}   & 100 & 68 & 32 \\
RoboVQA~\cite{sermanet2024}   & 100  & 93 & 7 \\
\bottomrule
\end{tabular}%
}
\end{table}

\begin{figure}[htbp]
    \centering
    \includegraphics[width=0.7\linewidth]{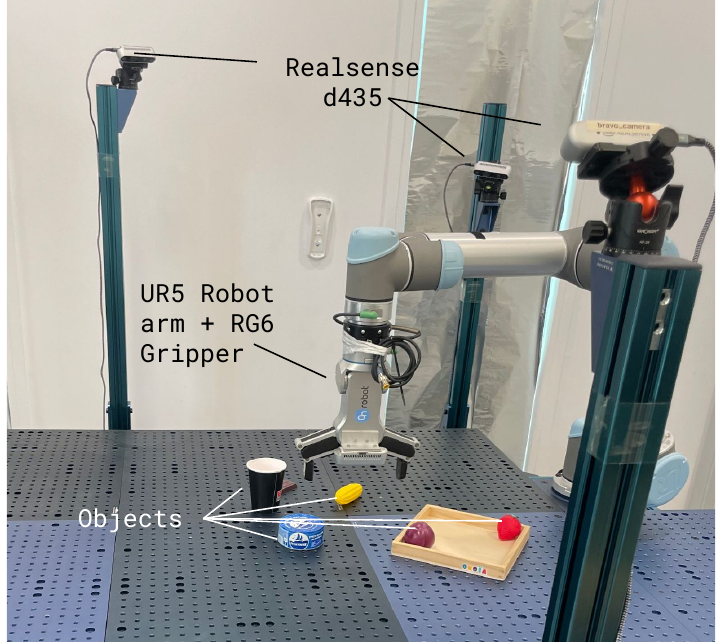}
    \caption{Our real-robot setup includes a UR5 robotic arm equipped with a RG6 gripper, with three RealSense D435 cameras.}
    \label{fig:ur5_realrobot_setup}
\end{figure}

\section{Real-robot Setup}
\label{sec:supp-mat-real-robot}

\cref{fig:ur5_realrobot_setup} shows our real-robot setup, we use a 6-DoF UR5 equipped with a RG6 gripper, on a tabletop scene. The setup contains multiple views with three RealSense D435 cameras.

\begin{table}[t]
\centering
\caption{Training and inference hyperparameters.}
\label{tab:supp_mat_hyperparameters}
\setlength{\tabcolsep}{4pt}
\renewcommand{\arraystretch}{1.05}
\resizebox{\linewidth}{!}{%
\begin{tabular}{@{}ll@{}}
\toprule
\multicolumn{2}{c}{\textbf{Training}} \\
\midrule
Base model & InternVL3-8B (InternViT-300M-448px-V2.5 + Qwen2.5-7B) \\
Loss & Next-token prediction \\
Optimizer & AdamW \\
Learning rate & $4 \times 10^{-5}$ \\
LR schedule & Cosine, warmup ratio 0.03 \\
Weight decay & 0.05 \\
LoRA rank & 16 \\
LoRA $\alpha$ & 32 \\
LoRA dropout & 0.05 \\
Frozen modules & Visual encoder, MLP projector, base LLM weights \\
Precision & bf16 \\
Effective batch size & 16 \\
Epochs & 5 \\
Max sequence length & 8192 \\
Image size & $448 \times 448$ \\
Distributed training & DeepSpeed ZeRO-1, gradient checkpointing \\
Hardware & H100 \\
\midrule
\multicolumn{2}{c}{\textbf{Inference}} \\
\midrule
Temperature & 0 \\
\bottomrule
\end{tabular}%
}
\end{table}
\section{Implementation Details}
\label{sec:supp-mat-implementation-details}

\cref{tab:supp_mat_hyperparameters} details the training and inference hyperparameters for Guardian. In addition to these, we release datasets, code, and models.

\end{document}